# Mixture Approximations to Bayesian Networks

**Volker Tresp, Michael Haft and Reimar Hofmann**
Siemens AG, Corporate Technology
Neural Computation
Dept. Information and Communications
Otto-Hahn-Ring 6, 81730 Munich, Germany

## Abstract

Structure and parameters in a Bayesian network uniquely specify the probability distribution of the modeled domain. The locality of both structure and probabilistic information are the great benefits of Bayesian networks and require the modeler to only specify local information. On the other hand this locality of information might prevent the modeler —and even more any other person— from obtaining a general overview of the important relationships within the domain. The goal of the work presented in this paper is to provide an "alternative" view on the knowledge encoded in a Bayesian network which might sometimes be very helpful for providing insights into the underlying domain. The basic idea is to calculate a mixture approximation to the probability distribution represented by the Bayesian network. The mixture component densities can be thought of as representing typical scenarios implied by the Bayesian model, providing intuition about the basic relationships. As an additional benefit, performing inference in the approximate model is very simple and intuitive and can provide additional insights. The computational complexity for the calculation of the mixture approximations critically depends on the measure which defines the distance between the probability distribution represented by the Bayesian network and the approximate distribution. Both the KL-divergence and the backward KL-divergence lead to inefficient algorithms. Incidentally, the latter is used in recent work on mixtures of mean field solutions to which the work presented here is closely related. We show, however, that using a mean squared error cost function leads to update equations which can be solved using the junction tree algorithm. We conclude that the mean squared error cost function can be used for Bayesian networks in which inference based on the junction tree is tractable. For large networks, however, one may have to rely on mean field approximations.

## 1 Introduction

One of the appealing aspects of Bayesian networks is their modularity; the modeler has to specify only local information and may thus generate a complex model step by step. As a drawback of local modeling one might very soon loose the overview of the relationships in the joint model. The goal of the work presented in this paper is to provide an "alternative" view on the domain knowledge encoded in a Bayesian network which might sometimes be very helpful for providing insights into the underlying domain. The basic idea is to calculate a mixture approximation to the probability distribution represented by the Bayesian network. The mixture component densities can be thought of as representing typical scenarios as components of the joint distribution thus providing intuition about the basic relationships in the domain. It can be argued that reasoning in terms of cases or scenarios is very close to the human approach to dealing with complex domains. This idea was explored by Druzdzel and Henrion (1990) who used as scenarios the most likely configurations given some evidence (although they didn't provide efficient algorithms for finding those). A mixture of 'scenarios' can be computed both for the joint distribution encoded by the Bayesian net and for a conditional distribution given evidence. In the latter case our approach can illustrate the relationships in the joint distribution of the unknown variables given the evidence which is not straightforward in the standard evidence propagation algorithms which primarily only calculate simple marginals. The challenge in our approach lies in calculating the optimal parameters in the mixture models. It turns out that the computational complexity for the calculation of the mixture approximations critically depends on the measure which is used for calculating the distance between the probability distribution represented by the Bayesian network and the approximate distribution. We show that finding the best approximation using the most natural metric, i.e. the Kullback-Leibler (KL-) distance is computationally infeasible. The same is true if the backward KL-distance is used. Haft, Hofmann and Tresp (1997, 1999) have shown that for the latter case, a reasonable approximation can be found by first finding the local minima of the backward KL-distance between the Bayesian network model and *one* mixture component (which corresponds to solving the mean field equations) and by forming the mixture model in a second step by using a small overlap assumption. A new approach pursued in this paper is to replace the KL-distance as a distance measure by the mean squared distance between the probability distribution described by the Bayesian network model and the mixture approximation. We can show that if either the mean squared distance or the expected mean squared distance is used, the resulting equations can be solved using the junction tree algorithm (Lauritzen and Spiegelhalter, 1988, Jensen, Lauritzen and Olsen, 1990). The algorithms are therefore exactly then efficient, when inference in the Bayesian network itself is efficient. The ideas here are developed for binary Bayesian networks but can easily be generalized to general graphical models, discrete models and Gaussian graphical models. In the next Section we develop and discuss the different algorithms for obtaining mixture approximations in the joint probability distribution. In the following Section we show how evidence can be taken into account. In Section 4 we perform a mixture analysis of the well-known chest clinic example. We demonstrate how typical scenarios can be extracted, how inference can be performed in the approximate model and how simple probabilistic rules can be extracted. Finally, in Section 5 we present conclusions.

## 2   Theory

Assume a Bayesian network with $N$ variables with probability distribution $P(x)$ which factorizes as

$$P(x) = \prod_{j=1}^{N} P(x_j | \Pi_j).$$

As stated in the introduction our goal is to find a mixture approximation to $P(x)$ of the form

$$P(x) \approx Q(x) = \sum_{i=1}^{M} Q(i, x) = \sum_{i=1}^{M} q_i \, Q(x|i). \quad (1)$$

For simplicity, we focus in this paper on the case that $x = (x_1, \ldots, x_N)'$ is a vector of binary variables $x_j \in \{0, 1\}$ and that the mixture component distributions factorize into binomial distributions

$$Q(x|i) = \prod_{j=1}^{N} q_{ij}^{x_j} (1 - q_{ij})^{1-x_j}. \quad (2)$$

The goal is now to determine the model parameters $\{q_{ij}\}_{i=1, j=1}^{M, N}$ and $\{q_i\}_{i=1}^{M}$ such that we obtain a good approximation to $P(x)$. First, we need to define a distance measure which specifies what exactly we mean by a "good" approximation. In the next sections we will define different distance measures and we will show that the complexity of the resulting update equations very much depends on the distance measure which is selected.

### 2.1   KL-divergence

The KL-divergence has the form

$$KL(P(x) \| Q(x)) = -\sum_{x} P(x) \, \log \frac{Q(x)}{P(x)}.$$

This cost function might be considered the most natural cost function since it corresponds to drawing an infinite number of samples from $P(x)$ and to then do a maximum likelihood mixture modeling approach of the data.

The $KL$-distance can be minimized using an EM-algorithm. The E-step calculates $\forall i$ and $\forall$ configurations of $x$

$$Q(i|x) = \frac{Q(x|i) q_i}{\sum_{i=1}^{M} Q(x|i) q_i}$$

and the M-steps update $\forall i, j$

$$q_{ij} = \frac{\sum_{x, x_j=1} P(x) Q(i|x)}{\sum_{x} P(x) Q(i|x)}$$

and $\forall i$

$$q_i = \frac{\sum_{x} P(x) Q(i|x)}{\sum_{x} \sum_{i}^{M} P(x) Q(i|x)}.$$

Since the summations in the update equations are over exponentially many states of $x$ and since $Q(i|x)$ cannot be easily decomposed, the update equations are infeasible for large networks. The only exception is if $M = 1$, since then the M-step reduces to $q_{1j} = P(x_j)$ and simply calculates the marginal distributions which can be calculated efficiently using the junction tree algorithm.

A simple approximate solution can be obtained by generating a large number of samples from a given Bayesian network and by then calculating the model which maximizes the likelihood w.r.t. the data using the corresponding EM-algorithm. In this case, the sums in the previous EM-equations contain only as many terms as data are generated. Although this approach should be feasible in most cases, if some mixture components only obtain small probabilities, one might have to generate a large number of samples to obtain good parameter estimates.

## 2.2 The Backward KL-divergence

There is a clear asymmetry in the KL-distance w.r.t the two distributions $P(x)$ and $Q(x)$. It is therefore also common to optimize the "backward" KL-distance defined as

$$BKL(P(x)||Q(x)) = KL(Q(x)||P(x))$$

$$= -\sum_x Q(x) \log \frac{P(x)}{Q(x)}.$$

Note that the role of $Q(x)$ and $P(x)$ is interchanged and the expectation is calculated with respect to the simpler approximate distribution $Q(x)$. For $M = 1$ (i.e. only one component), minima of $BKL$ can be found by iterating the associated mean field equations which for the binary case (2) read

$$q_{1,j} = sig\left(\sum_{M_j} Q(M_j) \, \log\left[\frac{P(x_j=1|M_j)}{1-P(x_j=1|M_j)}\right]\right).$$

The previous equation is iterated repeatedly for all $j$ until convergence where $M_j \subset x$ denotes the elements in the Markov blanket of $x_j$ and

$$sig(x) = [1 + \exp(-x)]^{-1}.$$

The previous update equation is efficient since it requires the summation only over the elements in the Markov blanket of $x_j$ and therefore only involves local computations. Haft, Hofmann and Tresp (1999) have shown that for Bayesian networks the update equations can further be simplified. Note, that –as for example in the mean field approximation to the Boltzmann machine— the sigmoid transfer function is a *result* of using the mean field approximation and was not postulated as in the work on sigmoid belief networks (Saul, Jaakkola and Jordan, 1996).

Although the previous mean field update equations can be used to find local optima in the $BKL$-distance, calculating the optimal best mixture approximation— as for the KL-distance— involves the summation over all states of the Bayesian network. Haft, Hofmann and Tresp (1999) have therefore suggested to find $M$ local optima of the BKL-distance and use those as components in the mixture model. The mixture weights can then be calculated using a small overlap assumption as

$$q_i = \frac{1}{C} \exp\left(-\sum_{j=1}^{M} \sum_{x_j, \Pi_j} Q(x_j, \Pi_j|i) \log \frac{Q(x_j|i)}{P(x_j|\Pi_j)}\right)$$

where $C$ normalizes the $q_j$ and $Q(x_j|i) = q_{ij}^{x_j}(1-q_{ij})^{1-x_j}$. Note, that again only local computations are required.

Further approximate solutions have been derived by Bishop, Lawrence, Jaakkola and Jordan (1998), and Lawrence, Bishop and Jordan (1998) for special graphical models such as Boltzmann machines and sigmoid belief networks.

## 2.3 The Mean Squared Error

Let's consider the squared error cost function

$$SE(P(x)||Q(x)) = \sum_x (P(x) - Q(x))^2$$

$$= \sum_x P(x)^2 + \sum_{i=1}^{M} \sum_{k=1}^{M} \sum_x q_i q_k Q(x|i) Q(x|k)$$

$$-2 \sum_{i=1}^{M} \sum_x P(x) \, q_i Q(x|i). \qquad (3)$$

The advantage now is that the cost function is a simple quadratic expression in the the parameters of the approximating mixture distribution (1). The SE cost function can be motivated by considering a Taylor expansion to the KL-distance which yields as a distance measure $\sum_x (P(x) - Q(x))^2 / P(x)$. By taking a closer look at Equation 3 we notice that all sums over $x$ have the form

$$\sum_x P(x)^A \prod_{j=1}^{N} f_j(x_j) \qquad (4)$$

where $A$ is an integer and $f_j$ is a function of $x_j$ only,, $\forall j$.

We make the following observation:

**Observation 1** *All summations of the form of Equation 4 can be calculated efficiently using, e.g., the junction tree algorithm, as long as the distribution $P$ itself can be handled efficiently.*

To see this it is sufficient to note that the structure of the sum (4) is still that of the original Bayesian network and therefore can be factorized the same way. To be explicit,

$$\sum_x P(x)^A \prod_{j=1}^N f_j(x_j) = \sum_x \prod_{j=1}^N P(x_j|\Pi_j)^A f_j(x_j). \quad (5)$$

Formally, $f_j(x_j)$ assumes the role of soft evidence in a Bayesian network defined by the potentially unnormalized conditional probabilities $P(x_j|\Pi_j)^A$ (Appendix 6.1).

We can calculate update equations by setting the derivatives of the cost function with respect to the parameters to zero. We obtain

$$q_{ij} = \frac{1}{2q_i^2 \sum_{x\setminus x_j} Q(i, x\setminus x_j)^2} \times \quad (6)$$

$$\left(\sum_x (2x_j - 1)P(x)Q(i, x\setminus x_j) + \sum_{x\setminus x_j} Q(i, x\setminus x_j)^2\right.$$

$$\left. + \sum_{l,l\neq i}(1 - 2q_{lj}) \sum_{x\setminus x_j} Q(x_j|l)\, Q(i, x\setminus x_j)\, Q(j, x\setminus x_j)\right)$$

where $Q(i, x\setminus x_j) = q_i \prod_{k=1, k\neq j}^N q_{ik}^{x_k}(1-q_{ik})^{1-x_k}$. All summations can be calculated efficiently by using Observation 1. Updating the parameters using Equation 6 performs component wise minimization of the cost function.

The optimization of the component weights $q_i$ is a low dimensional quadratic optimization problem with constraints $q_i > 0$ and $\sum_{i=1}^M q_i = 1$. The gradient of $SE$ with respect to a mixture component can also be calculated efficiently using Observation 1.

Note that since we can evaluate the distance measure $SE$ efficiently, we can easily decide if we supplied a sufficient number of mixture components for obtaining a good approximation.

### 2.4 The Expected Mean Squared Error

Alternatively, we might consider the expected mean squared error

$$ESE(P(x)\|Q(x)) = \sum_x P(x)(P(x) - Q(x))^2.$$

Similar as in the previous section, we obtain the fixed point equations for the $q_{ij}$,

$$q_{ij} = \frac{1}{\sum_x P(x)Q(i, x\setminus x_j)^2} \times \quad (7)$$

$$\left(\sum_x (2x_j - 1)P(x)^2 Q(i, x\setminus x_j)^2 + \sum_{x, x_j=0} P(x)Q(i, x\setminus x_j)^2\right.$$

$$\left. + \sum_{l,l\neq i} \sum_x (1 - 2x_j)P(x)\, Q(x_j|l)\, Q(i, x\setminus x_j)\, Q(l, x\setminus x_j)\right)$$

All sums can be calculated efficiently using Observation 1.

## 3 Evidence and Inference

If evidence is entered into the Bayesian network we can obtain a mixture approximation to the conditional distribution of the remaining variables given the evidence in the same way as described in the previous section. The mixture distribution for the unknown nodes may then be viewed as defining possible scenarios given our state of information.

Having already computed a mixture approximation to the joint distribution there is of course a much simpler way to obtain a mixture model for the conditional distribution. Rather than propagating the evidence in the exact model and then reapproximating the conditional distribution with new mixture components and mixing weights, one can perform approximate inference in the mixture model directly:

$$Q(u|e) = \frac{\sum_{i=1}^M q_i\, Q(e|i) Q(u|i)}{\sum_{i=1}^M q_i\, Q(e|i)}$$

with

$$Q(e|i) = \prod_{x_j \in e} q_{ij}^{x_j}(1 - q_{ij})^{1-x_j}$$

and

$$Q(u|i) = \prod_{x_j \in u} q_{ij}^{x_j}(1 - q_{ij})^{1-x_j}$$

where $e \subset x$ are the variables with evidence and $u = x\setminus e$ are the remaining variables. The above equation shows that the conditional distribution $Q(u|e)$ is still a mixture distribution composed of the same 'unconditional scenarios', but the new mixture weights are proportional to

$$q_i Q(e|i).$$

That is, the effect of evidence is just a reweighting of the known scenarios, which makes the effects of the evidence clearly visible. Each mixture component is weighted by the product of $q_i$ (as before) and the probability of the evidence $Q(e|i)$ for that mixture component.

## 4 Experiments

### 4.1 Modeling

The goal of our experiments was to test the quality of the mixture solutions found by the different cost functions and algorithms. As a test case we used the well known chest clinic example. The chest clinic is a well documented Bayesian network (Lauritzen and Spiegelhalter, 1988) consisting of the 8 variables *visit to asia* (1), *smoker* (2), *tuberculosis* (3), *lung cancer* (4), *bronchitis* (5), *tuberculosis or lung cancer* (6), *positive x-rays* (7) and *dyspnoea* (8). In this domain with only eight variables the algorithms for all cost functions could be executed in reasonable time. The network structure is defined in Appendix 6.2.

In the first experiment the EM-algorithm using the KL-distance measure as described in Section 2.1 was used. Figure 1 shows the KL-distance between the true probability distribution and the approximate distribution using different number of components. A small KL-divergence is reached for only 4 mixture components. If more components are used, the description is further refined but the KL-divergence does not improve significantly.

In the next experiment, the mixture parameters were calculated by minimizing the KL-distance and the SE and ESE distance measures using four and five mixture components. Table 1 and 2 show that for four or five mixture components only three mixture components obtain weights of more than 1% for all approaches.

Figure 2 and 3 plot $q_{ij}$ (the probability for a positive finding of node $j$ in scenario $i$) for the solutions obtained by the three algorithms. Tables 1 and 2 show the mixture weights and the KL-distance of the mixture approximations to the true probability distribution. We see that approximately half of the weight is assigned to a scenario describing healthy patients with a very low probability of any symptom and with approximately 30% probability of being a smoker. This mixture component really models two scenarios: one in which the healthy patient is a smoker and one where he or she is not a smoker. Approximately 40% probability mass is assigned to a group of patients with very high probability of bronchitis, high probability of dyspnoea and above average probability of being a smoker. Similarly as for the healthy patient, this mixture component can be thought of as modeling four patient groups all of which have bronchitis but vary in the four possible configurations of being smokers and having dyspnoea.[1] Approximately 5% probability mass is assigned to a group of patients with a very high probability of a positive x-ray, dyspnoea, a very high probability of lung cancer and a high probability of having bronchitis. In the model with five mixture components about 1% or less probability mass is assigned to a group of persons with tuberculosis, slightly higher than normal probability of having visited asia, and with a high probability of having a positive x-ray and dyspnoea. In the four component model, both the KL-distance model and the SE model have also converged to this solution.

---

[1] Note, that given the states of the remaining variables in this configuration, smoker and dyspnoea are likely to be independent; otherwise the mixture approximation would have had the tendency to assign more than one component to the bronchitis scenario.

Table 1: KL-divergence and component weights for the four component model.

| Approach | KL-dist | q1 | q2 | q3 | q4 |
|---|---|---|---|---|---|
| KL  | 0.0021 | 0.530 | 0.405 | 0.055 | 0.010 |
| SE  | 0.0055 | 0.522 | 0.415 | 0.053 | 0.001 |
| ESE | 0.1090 | 0.516 | 0.415 | 0.052 | 0.004 |

Interestingly, the mixture components really "make up their mind" and converge to clearly identifiable scenarios. The solutions from the different approaches all agree in the scenarios which have obtained considerable probability mass. In terms of KL-divergence (recall that KL-divergence is only minimized for one approach) the SE solution performs very well whereas the ESE-solution is considerably worse. This might be explained by the fact that ESE does not tend to penalize errors for states with small probabilities.

Based on these experiments we might conclude that the SE-approach can provide solutions very similar both qualitatively and quantitatively to the KL-solution.

Figure 4 displays the results using the mixture of mean field solutions approach (Haft, Hofmann and Tresp, 1999). First of all it is interesting to note that there are exactly three optima of the mean field equations. Approximately 91% of the probability mass is here assigned to the "normal" patient group with an average probability for smoking and bronchitis. The main difference to the first scenario of Figure 2 is that here the probability for bronchitis is higher. The second component with approximately 7% of probability mass is almost identical to the third component in the previous solutions. Finally, the third solution with approximately 1% of probability mass is very similar to the fourth component in the previous five component models. It appears that the main difference to the previous solutions is that the first component of the mean field solution approximates the first two components of the mixture approximations in the previous experiments.

### 4.2 Evidence and Inference

When we enter evidence as described in Section 3 the scenarios are correspondingly re-weighted. Table 3 describes the new mixture weights when the evidence

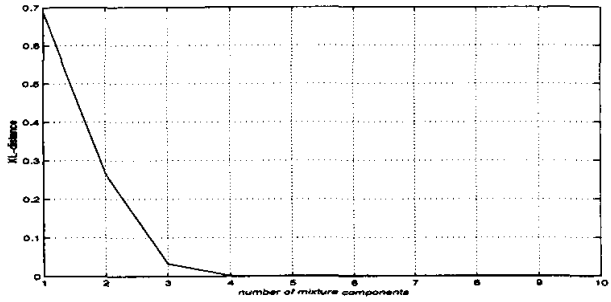

Figure 1: The mixture approximation using the KL-distance. The x-axis indicates the number of mixture components and the y-axis the KL-distance between the mixture model and the Bayesian network.

Table 2: KL-divergence and component weights for the five component model.

| Appr. | KL-dist | q1 | q2 | q3 | q4 | q5 |
|---|---|---|---|---|---|---|
| KL | 0.0020 | 0.517 | 0.404 | 0.055 | 0.010 | 0.014 |
| SE | 0.0056 | 0.522 | 0.415 | 0.043 | 0.009 | 0.012 |
| ESE | 0.0360 | 0.513 | 0.422 | 0.052 | 0.005 | 0.008 |

"dyspnoea=yes" and "smoker=yes" is entered. One can see, for example, that for the SE metric the first scenario (healthy person) has lost and the second (person with bronchitis) and third (sick smoker) scenarios have gained weight as a result of the evidence — if compared to unconditional weights in Table 1. Table 4 shows the marginal posterior probabilities computed from the different mixture models. As could be expected from the low KL-distances between the exact and all mixture models, the estimated probabilities are very close to the true probabilities. The KL-metric shows the best and the ESE-metric the worst approximation, but all three are within around one percentage point from the exact posteriors for all variables.

Table 5 and Table 6 show the same experiment for the *unlikely* evidence "visit to Asia=yes" and "x-ray=positive" which has a probability of only 0.15% in the model. Even though the KL-distance between the exact (unconditioned) model and the mixture models is low, the KL-distance of the corresponding conditioned models can be large, because a small absolute error in the estimated probability of an unlikely event causes only a small KL-distance, but a large relative error in the probability of that event. This, however, can lead to a large absolute error in the probabilities conditioned on that event. We therefore expect stronger deviations in the probabilities for this second example. Table 6 shows indeed stronger deviations than Table 4. The KL-model is still extremely good, the SE-model shows deviations in the range of 5% percentage points, whereas the ESE-model is completely wrong in its es-timation of the probability of TBC. This is probably due to the fact that deviations from the exact model in regions of low probability have a particularly low weight in the ESE-metric. So both the KL- and the SE-model still give useful results even in this case of unlikely evidence.

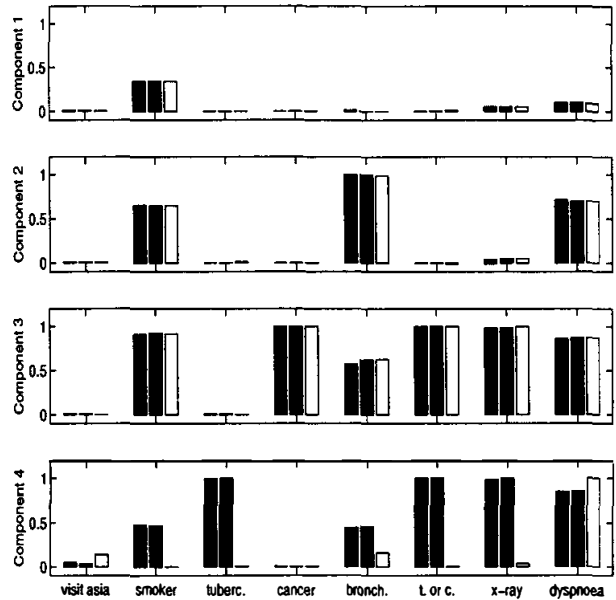

Figure 2: The component parameters $q_{ij}$ for the mixture model with four components $i = 1, 2, 3, 4$. The filled bars show the result using the KL-divergence, the gray bars using the SE error function, and the empty bars using the ESE-error function.

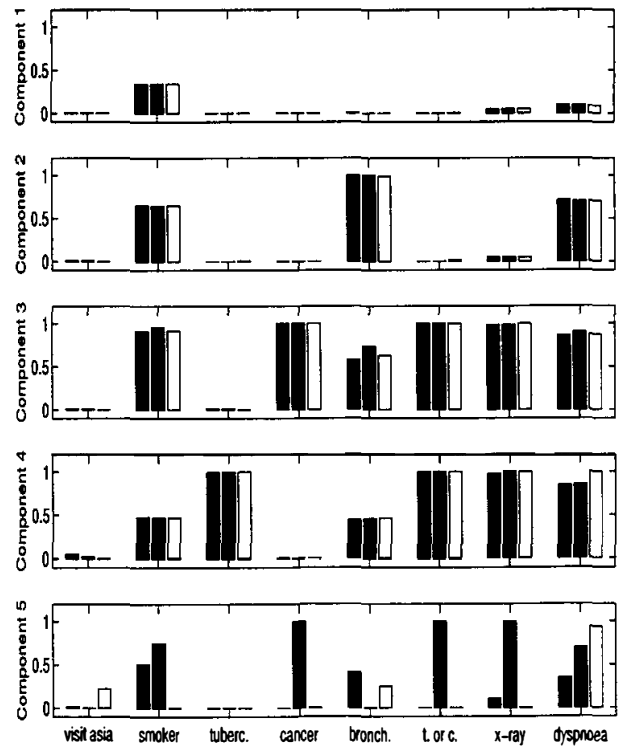

Figure 3: Same as in Figure 2 but with five components.

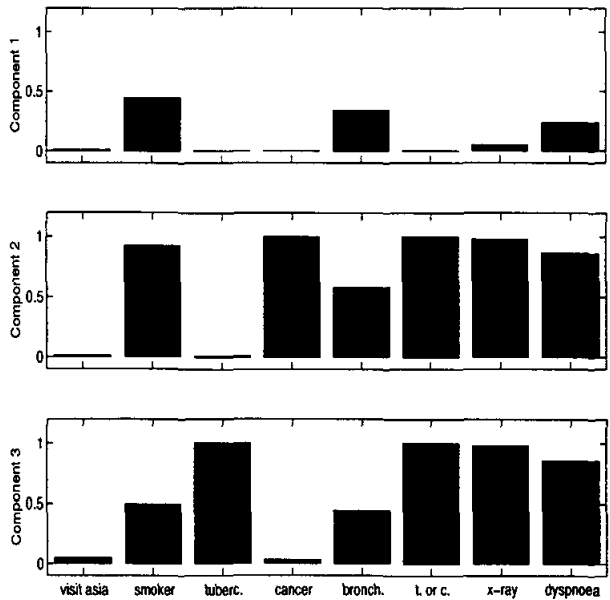

Figure 4: The solution obtained by the mean field approximation. There are exactly three solutions. The mixture weights are 0.919 (component 1), 0.069 (component 2) and 0.012 (component 3). The KL-distance between the true probability distribution and the mixture approximation is 0.304.

Table 3: Reweighted mixture weights for evidence "dyspnoea=yes" and "smoker=yes".

| Appr. | q1 | q2 | q3 | q4 |
|---|---|---|---|---|
| KL | 0.0715 | 0.7434 | 0.1692 | 0.0159 |
| SE | 0.0705 | 0.7445 | 0.1704 | 0.0147 |
| ESE | 0.0621 | 0.7724 | 0.1655 | 0.0000 |

## 5 Conclusions

We have demonstrated that a mixture approximation can give interesting new insights into the relationships between the variables in a probability domain modeled by a Bayesian network. Furthermore, we have shown that the complexity of the computational cost for calculating the parameters in the mixture models depends critically on the distance measure. For a quadratic distance measure, parameter optimization can be executed efficiently using the junction tree algorithm. We have show that experimentally, the so-lutions found by minimizing the KL-distance (which is computationally infeasible for large networks) and a squared error cost function are basically identical. The significant components in each mixture model clearly make up their mind and are very different. We have compared these results to those obtained using multiple mean field solutions. Here, maxima in the backward KL-distance are found and form the mixture components. The mean field solutions have been used recently for approximate inference and learning in various large intractable networks, for example by Peterson and Anderson (1987), Saul, Jaakkola and Jordan (1996), Kappen and Rodriguez (1997), Bishop, Lawrence, Jaakkola and Jordan (1998), Lawrence, Bishop and Jordan (1998) and Haft, Hofmann and Tresp (1999). For those large networks in which the junction tree algorithm is computationally infeasible, mean field approximations are still the only viable option for inference, besides Monte Carlo methods. For smaller networks in which the junction tree algorithm can be used, the algorithms presented here are applicable. Finally, we have shown how the mixture approximation can be used to obtain both a mixture model for the unknown nodes composed of reweighted scenarios and approximations to conditional probabilities. The inference based on the mixture approximation appears to be reasonable precise, as long as the inference does not depend on a very accurate model of the underlying probability distribution; the latter is the case when the evidence entered into the model is very unlikely.

Table 5: Reweighted mixture weights for the (unlikely) evidence "visit to Asia=yes" and "x-ray=positive".

| Appr. | q1 | q2 | q3 | q4 |
|---|---|---|---|---|
| KL | 0.1749 | 0.1338 | 0.3670 | 0.3242 |
| SE | 0.2127 | 0.1650 | 0.3595 | 0.2627 |
| ESE | 0.2321 | 0.4921 | 0.2045 | 0.0714 |

### Acknowledgements

The comments of the three anonymous reviewers were very helpful.

### References

[1] C. M. Bishop, N. D. Lawrence, T. Jaakkola, and M. I. Jordan. Approximating posterior distributions in belief networks using mixtures. In M. I. Jordan, M. J. Kearns, and S. A. Solla, editors, *Advances in Neural Information Processing Systems 10*, pages 416–422. MIT Press, Cambridge MA, 1998.

[2] J. Druzdzel, M. and M. Henrion. Using scenarios to explain probabilistic inference. In *Workshop*

Table 4: Inference based on the various mixture models given the evidence "dyspnoea=yes" and "smoker=yes". Shown is also the result of exact inference.

| Appr. | Asia | Smoker | TBC | Lunc Cancer | Bronchitis | TBC or LC | X-Ray | Dyspnoea |
|---|---|---|---|---|---|---|---|---|
| KL | 0.0103 | 1.0000 | 0.0174 | 0.1693 | 0.8484 | 0.1851 | 0.2222 | 1.0000 |
| SE | 0.0094 | 1.0000 | 0.0169 | 0.1710 | 0.8563 | 0.1857 | 0.2225 | 1.0000 |
| ESE | 0.0049 | 1.0000 | 0.0101 | 0.1662 | 0.8663 | 0.1760 | 0.2068 | 1.0000 |
| exact | 0.0103 | 1.0 | 0.0178 | 0.1707 | 0.8598 | 0.1867 | 0.2236 | 1.0 |

Table 6: Inference based on the various mixture models given the (unlikely) evidence "visit to asia=yes" and "x-ray=positive". Shown is also the result of exact inference.

| Appr. | Asia | Smoker | TBC | Lunc Cancer | Bronchitis | TBC or LC | X-Ray | Dyspnoea |
|---|---|---|---|---|---|---|---|---|
| KL | 1.0000 | 0.6352 | 0.3248 | 0.3682 | 0.4905 | 0.6913 | 1.0000 | 0.7013 |
| SE | 1.0000 | 0.6348 | 0.2658 | 0.3598 | 0.5082 | 0.6220 | 1.0000 | 0.6730 |
| ESE | 1.0000 | 0.5843 | 0.0072 | 0.2051 | 0.6256 | 0.2131 | 1.0000 | 0.6144 |
| exact | 1.0 | 0.6370 | 0.3377 | 0.3715 | 0.4911 | 0.6906 | 1.0 | 0.7011 |

on Explanation, pages 133–141. American Association for Artificial Intelligence, 1990.

[3] H. Haft, R. Hofmann, and V. Tresp. Model-independent mean-field theory as a local method for approximate propagation of information. Network, 10(1):93–105, 1999.

[4] M. Haft, R. Hofmann, and V. Tresp. Model-independent mean field theory as a local method for approximate propagation of information. Technical Report, http://www7.informatik.tu-muenchen.de/~hofmannr/mf_abstr.html, 1997.

[5] F. V. Jensen, S. L. Lauritzen, and K. G. Olsen. Bayesian updating in causal probabilistic networks by local computations. Computational Statistics Quaterly, 4:269–282, 1990.

[6] H. J. Kappen and F. B. Rodriguez. Mean field approach to learning in boltzmann machines. Pattern Recogntion Letters, 18:1317–1322, 1997.

[7] S. L. Lauritzen and D. J. Spiegelhalter. Local computations with probabilities on graphical structures and their application to expert systems. J. Roy. Statist. Soc. B, 50:154–227, 1988.

[8] N. D. Lawrence, C. M. Bishop, and M. I. Jordan. Mixture representations for inference and learning in boltzmann machines. In G. F. Cooper and S. Moral, editors, Uncertainty in Artificial Intelligence: Proceedings of the Fourteenth Conference, pages 320–327. Morgan Kaufmann, San Francisco, CA, 1998.

[9] C. Peterson and J. R. Anderson. A mean field theory learning algorithm for neural networks. Complex Systems, 1:995–1019, 1987.

[10] K. L. Saul, T. Jaakkola, and M. I. Jordan. Mean field theory for sigmoid belief networks. Journal of Artificial Intelligence Research, 4:61–76, 1996.

## 6 Appendix

### 6.1 Interpretation of $f_j(x_j)$ as Soft Evidence

Consider the last term in Equation 3. If network we add soft evidence nodes $e_j$ with conditional densities $P(e_j|x_j) \propto q_i Q(x_j|i)$ to the original we obtain for the joint probability distribution $P(x) \prod_j P(e_j|x_j)$. The likelihood of the evidence is

$$P(e) = \sum_x P(x) \prod_j P(e_j|x_j)$$

and can be calculated efficiently using the junction tree algorithm.

### 6.2 The Chest Clinic

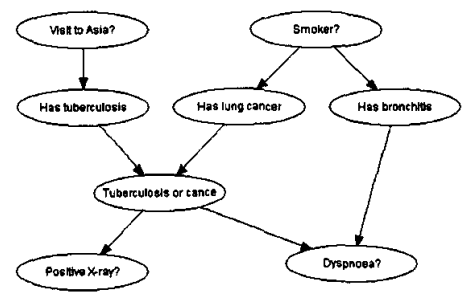

Figure 5: The chest clinic network. The parameters for this standard network can be downloaded, e.g., form http://www.hugin.dk/networks.